\documentclass[runningheads,a4paper]{llncs}

\usepackage{amssymb}
\usepackage{amssymb}
\usepackage{amsmath}
\usepackage{mathtools}
\usepackage{multirow}
\usepackage{subfig}
\usepackage{bm}
\usepackage{color, soul}
\usepackage[font={footnotesize}]{caption}
\usepackage{array}
\usepackage{paralist}
\usepackage[dvipsnames]{xcolor}
\usepackage[inline]{enumitem}
\usepackage{url}
\usepackage[draft]{pgf}
\usepackage{todonotes}

\urldef{\mail}\path|n.strisciuglio@rug.nl |
\newcommand{\keywords}[1]{\par\addvspace\baselineskip
\noindent\keywordname\enspace\ignorespaces#1}
\DeclareMathOperator*{\argmaxB}{argmax}  
\newcolumntype{C}[1]{>{\centering\let\newline\\\arraybackslash\hspace{0pt}}m{#1}}

\DeclareSymbolFont{extraup}{U}{zavm}{m}{n}
\DeclareMathSymbol{\vardiamond}{\mathalpha}{extraup}{87}

\begin{document}
\sloppy
\mainmatter  

\title{Detection of curved lines with \textit{B}-COSFIRE filters: A case study on crack delineation}

\titlerunning{Lecture Notes in Computer Science: Authors' Instructions}

%

\author{Nicola Strisciuglio\inst{1}
\and George Azzopardi\inst{1,2} \and Nicolai Petkov\inst{1}}
\authorrunning{N. Strisciuglio \and G. Azzopardi \and N. Petkov}

\institute{Johann Bernoulli Institute for Mathematics and Computer Science,\\
University of Groningen, The Netherlands\\
\and
Intelligent Computer Systems, University of Malta, Malta\\
\mail}

%
%

\toctitle{Lecture Notes in Computer Science}
\tocauthor{Authors' Instructions}
\maketitle

\begin{abstract}
The detection of curvilinear structures is an important step for various computer vision applications, ranging from medical image analysis for segmentation of blood vessels, to remote sensing for the identification of roads and rivers, and to biometrics and robotics, among others. 
This is a nontrivial task especially for the detection of thin or incomplete curvilinear structures surrounded with noise. We propose a general purpose curvilinear structure detector that uses the brain-inspired trainable \textit{B}-COSFIRE filters. It consists of four main steps, namely nonlinear filtering with \textit{B}-COSFIRE, thinning with non-maximum suppression, hysteresis thresholding and morphological closing. We demonstrate its effectiveness on a data set of noisy images with cracked pavements, where we achieve state-of-the-art results (F-measure $=0.865$). The proposed method can be employed in any computer vision methodology that requires the delineation of curvilinear and elongated structures.

\end{abstract}

\graphicspath{ {./figures/}
				{./figures/method/}
				{./figures/experiments/}}				

\section{Introduction}
The detection of curvilinear and elongated structures is of great importance in image processing due to its application to numerous problems. 
The delineation of blood vessels in medical images, the detection and measure of cracks in walls and roads for damage estimation, the segmentation of river and roads in aerial and satellite images to prevent disasters or accidents are few applications of algorithms for the detection of curvilinear patterns. 
 
In the literature, various approaches for curvilinear structure detection were proposed, for which a survey was recently published in~\cite{Bibiloni2016}. Existing methodologies range from parametric methods to approaches based on filtering techniques or region growing, point processes and machine learning. 
For instance, the Hough transform is a parametric method, which converts an input image to a parameter space where line or circle segments can be detected. It requires a mathematical model of the patterns of interest. Different elongated structures, such as lines, circles and Y-junctions, require different mathematical models to transform an image into the particular parameter space where the patterns of interest can be distinguished.

Approaches based on filtering techniques employ local derivatives in a multi-scale analysis~\cite{Frangi1998} or model the profile of elongated structures by means of a two-dimensional Gaussian kernel~\cite{HooverStare2000}.
Region-growing  that considers multi-scale information about width, length and orientation of lines was proposed in~\cite{MartinezPerez}. In~\cite{Mendonca2006}, mathematical morphology and tracking techniques were combined with a-priori information about the line network of interest. 
These methods are intuitive but require a-priori structural information about the patterns of interest. 

Despite their high computational complexity, point and object processes were proposed to detect line networks in images, especially in applications of road and river detection in aerial images. Methods in this group are based on tracking  elongated structures by simulation of complex mathematical models. In~\cite{Lacoste2005}, a line network is considered as a set of interacting  line segments which are reconstructed by object processes. Point processes based on the Gibbs model and Monte Carlo simulations were introduced in~\cite{Lafarge2010} and~\cite{Verdi2012}, respectively. In~\cite{chai2013} and ~\cite{Turetken16}, point processes were combined with a graph-based representation and classification to improve the accuracy of segmentation. Point processes and graph-based approaches require high computational resources, consequently reducing the applicability to high resolution images.

In the last group, there are methods that employ machine learning techniques. They are based on the construction of pixel-wise feature vectors and the use of classifiers to decide whether a pixel is part of an elongated structure or not. In~\cite{Niemeijer2004}, the responses of multi-scale Gaussian filters were used in combination with a $k$-NN classifier, while in~\cite{StaalDrive2004} a feature vector was constructed with the responses of a bank of ridge detectors. In~\cite{Soares2006}, the coefficients of multi-scale Gabor wavelets were used to form a feature vector and to train a Bayesian classifier. Recently, a convolutional neural network trained with image patches of lines was proposed in~\cite{Liskowski2016}. These methods are more complex than filtering-based approaches and require long training time. Furthermore, the classifiers can be trained only when ground truth is available, which is not always possible or prohibitively expensive to obtain.

In this work, we present a method for the detection of curvilinear structures, composed of four steps: \begin{enumerate*}
\item \textit{B}-COSFIRE filtering,
\item thinning with non-maximum supression,
\item hysteresis thresholding and
\item morphological closing
\end{enumerate*}. The basic idea of the \textit{B}-COSFIRE filters, that we originally proposed for retinal vessel segmentation in~\cite{AzzopardiMEDIA2015,StrisciuglioVIP15}, is inspired by the functions of simple cells in area V1 of visual cortex selective to elongated patterns of certain widths and orientations. 

The \textit{B}-COSFIRE filter is trainable as its structure is not fixed in the implementation, but it is learned in an automatic configuration process performed on a prototype pattern. 
The concept of trainable filters was introduced in~\cite{AzzopardiPetkovCOSFIRE2013} and  employed in image analysis~\cite{AzzopardiPetkovCORF2012}, object recognition~\cite{Gecer2017} and adapted to audio analysis~\cite{StrisciuglioCOPE2016}.
The trainability of the COSFIRE approach concerns the learning of the structure of the filters directly from prototype patterns. This aspect can be considered a kind of \emph{representation learning}. Similarly to deep learning but considering a single training sample at time, it aims at avoiding a feature engineering process and building adaptive pattern recognition systems.



We perform experiments  on a publicly available data set, namely Crack\_PV14 data set~\cite{Zou16} and compare the resulting performance to those of existing methods. We show the effectiveness of the proposed method for the detection of curvilinear and elongated structures, the robustness of \textit{B}-COSFIRE filters to incomplete lines, noise and tortuosity, and their application in a pipeline for crack detection in pavement and road images. 

The paper is organized as follows. In Section~\ref{sec:method} we present the proposed curved line operator. In Section~\ref{sec:results} we describe the Crack\_PV14 data set and the experimental protocol that we followed, compare the results that we achieved with the ones reported in the literature and discuss certain aspects of the proposed method. Finally, we draw conclusions in Section~\ref{sec:conclusions}.
 
 \begin{figure}[!t]
	\centering
	\scriptsize
   \input{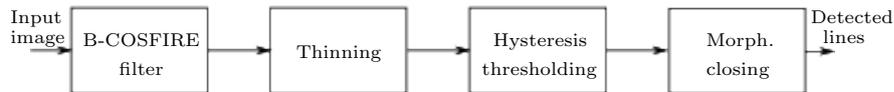}
   \caption{High-level architecture of the proposed curved line operator. A rotation-tolerant \textit{B}-COSFIRE filter responds to lines of preferred width and length. Thinning, hysteresis thresholding and  morphological closing are performed to obtain the final elongated and curvilinear binary structures.}
   \label{fig:arch}
\end{figure}

\section{Method}
\label{sec:method}
\subsection{Overview}
In Fig.~\ref{fig:arch}, we show the main steps of the proposed curved line detector. The  architecture for the detection of curvilinear and elongated patterns, which we apply to the detection of cracks in pavement and road images, is composed of four steps: 
\begin{enumerate*}
  \item \textit{B}-COSFIRE filtering,
  \item thinning with non-maximum suppression,
    \item hysteresis thresholding and
\item morphological closing.
\end{enumerate*}

Below we present the \textit{B}-COSFIRE filter and show how it is incorporated into the proposed curved line operator that is suitable for the detection of cracks in roads and walls.

\subsection{Configuration of a \textit{B}-COSFIRE filter}

A \textit{B}-COSFIRE filter takes input from 
of a group of Difference-of-Gaussians functions $DoG_\sigma$, with the outer Gaussian function having standard deviation  $\sigma$:
\begin{equation}
	DoG_\sigma(x,y) = \frac{\exp{\big(-\frac{x^2+y^2}{2(0.5\sigma)^2}\big)}}{2\pi(0.5\sigma)^2}-\frac{\exp{\big(-\frac{x^2+y^2}{2\sigma^2}\big)}}{2\pi\sigma^2}
\end{equation}
\noindent In \cite{nicolai2005modifications}, it was shown that for the above function the maximum response is elicited for a spot with a radius of $0.96\sigma$ or a line with a width of $1.92\sigma$. 
Based on this finding we set the outer standard deviation $\sigma = w/1.92$ where $w$ is the preferred width of the line of interest. 

The structure of a \textit{B}-COSFIRE filter, i.e. the positions at which we consider the DoG responses, is determined in an automatic configuration process on a given prototype pattern. For details about the configuration we refer the reader to~\cite{AzzopardiMEDIA2015}. We configure a \textit{B}-COSFIRE filter on a prototype line structure of width $w$, length $l$ and orientation $\phi$. 
The result of the configuration is a set $B_{w,l,\phi}$:
\begin{gather}
	B_{w,l,\phi} = \{(0,0),(\lambda,\phi),(\lambda,2\pi\!-\!\phi)\} \cup \{(\rho_i,\phi)\} \cup \{(\rho_i,2\pi\!-\!\phi)\} 	
\end{gather}
\noindent where $\lambda = \lfloor(l-1)/2\rfloor$ and $\rho_i = \eta i$ with $i = 1, \dots, \lfloor(\lambda-1)/\eta \rfloor-1$. $\phi$ is the preferred orientation of the line. The two-tuples in the set $B_{w,l,\phi}$ indicate the positions (distances and polar angles) with respect to the \textit{B}-COSFIRE filter support center at which we take the responses of a center-on difference-of-Gaussians (DoG) filter. The parameter $\eta$ (with $1 \le \eta \le \lambda \nonumber$) represents the pixel spacing between the considered DoG responses. When $\eta=1$ we configure a tuple for every location along a line with preferred width $w$, length $l$, and orientation $\phi$, and when $\eta = \lfloor(l-1)/2 \rfloor$ (i.e. the maximum possible value) the resulting filter consists of only three tuples: the tuple $(0,0)$ that refers to the DoG response at the center, and the two tuples $(\lfloor(l-1)/2\rfloor, \phi)$ and $(\lfloor(l-1)/2\rfloor,2\pi-\phi)$ that refer to the farthest distances on both sides of the support. The selectivity of a \textit{B}-COSFIRE filter increases with decreasing $\eta$ value.

\subsection{Response of a \textit{B}-COSFIRE filter}
The response of a \textit{B}-COSFIRE filter $B_{w,l,\phi}(x,y)$ is computed in four steps, namely \textit{filter-blur-shift-combine}. 

In the first step we \textit{filter} an input image $I$ with the DoG kernel $DoG_{\sigma=w/1.92}$ and denote the resulting image by $C$:

\begin{equation}
C(x,y) = \bigg|\sum_{x' = -3\sigma}^{3\sigma}\sum_{y' = -3\sigma}^{3\sigma} I(x,y)  DoG_\sigma(x-x',y-y')\bigg|^+
\end{equation}

\noindent where the operation $|.|^+$ denotes half-wave rectification, recently also known as rectifying linear unit (ReLU). 

Then, in order to allow for some tolerance with respect to the preferred positions we \textit{blur} the DoG responses 
by a nonlinear blurring operation that consists in a weighted maximum. The weighting is given by a Gaussian function whose standard deviation $\sigma'_i$ increases linearly with an increasing distance from the support center of the \textit{B}-COSFIRE filter: $\sigma'_i = \sigma'_0 + \alpha\rho_i$. The values $\sigma'_0$ and $\alpha$ are parameters and regulate the tolerance to deformations of the prototype pattern.

In the third step, we \textit{shift} the $i$-th blurred DoG response by a vector ($\rho_i, 2\pi-\phi_i$). In this way, all involved DoG responses meet at the same location, that is the support center of the concerned \textit{B}-COSFIRE filter. We denote by $s_{\rho_i,\phi_i}(x,y)$ the blurred and shifted DoG response at the location $(x,y)$ of the $i$-th tuple:

\begin{equation}
s_{\rho_i,\phi_i}(x,y) = \max_{x',y'}\{C(x-x'-\Delta x,y-y'-\Delta y)~G_{\sigma'_i}(x',y')\}
\end{equation}

\noindent where $\Delta x = -\rho_i \cos \phi_i$ and $\Delta y = -\rho_i \sin \phi_i$.

Finally, we denote by $r_{B_{w,l,\phi}}(x,y)$ the response of a \textit{B}-COSFIRE filter, which we compute by geometric mean:

\begin{equation}
r_{B_{w,l}}(x,y) = \bigg( \prod_{i=1}^{|B_{w,l}|} s_{\rho_i,\phi_i}(x,y)\bigg)^{1/ |B_{w,l}|}.
\end{equation}


\begin{figure}[!b]
\footnotesize
\centering
	\begin{tabular}{cccc}
		\includegraphics[height=27mm]{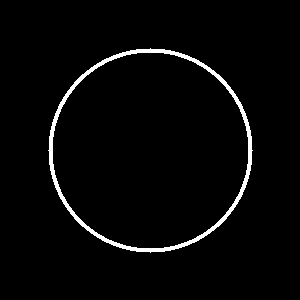}&
		\includegraphics[height=27mm]{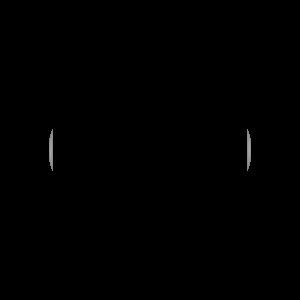}&
		\includegraphics[height=27mm]{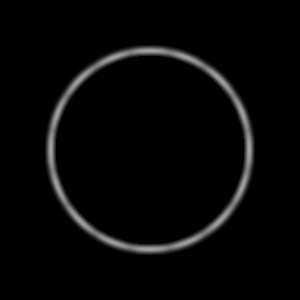}&
		\includegraphics[height=27mm]{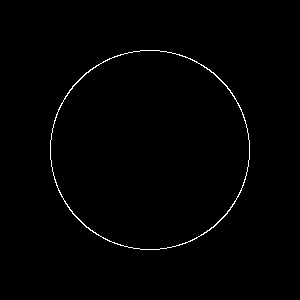}\\
		(a) & (b) & (c) & (d) 
    \end{tabular}
\caption{(a) An input image (of size $300 \times 300$ pixels) containing a circle with a circumference of radius $100$ pixels and a line width of $5$ pixels. (b) The  response map obtained by the \textit{B}-COSFIRE filter $B_{w=5,l=59,\phi=0}$. For $\sigma'_0=5$ and $\alpha=1$ the orientation bandwidth at $75\%$ of the maximum is circa $\pi/8$ radians. (c) The rotation-tolerant response map with eight orientations and (d) the thinned and binary image.}
\label{fig2}
\end{figure}

\subsection{Orientation bandwidth and tolerance to rotation}
The orientation bandwidth of a \textit{B}-COSFIRE filter is controlled by the parameters $\sigma'_0$ and $\alpha$. 
In the example of Fig.~\ref{fig2}, the \textit{B}-COSFIRE filter that is selective for lines of length 59 pixels achieves an orientation bandwidth (full width at 75\% of the maximum) of circa $\pi/8$ radians, for $\sigma'_0=5$ and $\alpha=1$.

We configure a set ${\beta = \{B_{w,l,\phi=\theta}\ \lvert ~\theta = 0, \pi/8, \dots, 7\pi/8 \} }$ of \textit{B}-COSFIRE filters  with eight orientation preferences. With a bandwidth of circa $\pi/8$ radians, a set of eight \textit{B}-COSFIRE filters with equidistant orientation preference is sufficient  to respond to lines in any orientation.
We denote by $\hat{r}_\beta(x,y)$ and $\Phi_\beta(x,y)$ the rotation-tolerant response and the orientation map of the rotation-tolerant \textit{B}-COSFIRE filter:
\begin{gather}
\hat{r}_\beta(x,y) = \max \{r_{B_{w,l,\phi=0}}(x,y), r_{B_{w,l,\phi=\pi/4}}(x,y), \dots, r_{B_{w,l,\phi=7\pi/8}}(x,y)\} \\
\Phi_\beta(x,y) = \argmaxB_{\phi} \{r_{B_{w,l,\phi=0}}(x,y), r_{B_{w,l,\phi=\pi/4}}(x,y), \dots, r_{B_{w,l,\phi=7\pi/8}}(x,y)\}
\end{gather}


\subsection{Binary map with thinning and hysteresis thresholding}
%

In order to obtain a binary map of curvilinear structures, we apply a thinning and hysteresis thresholding operations to the response map of the rotation-tolerant \textit{B}-COSFIRE filter. We use the thinning algorithm described in \cite{grigorescu2004contour} that takes as input the response map $\hat{r}_\beta(x,y)$ and the orientation map $\Phi_\beta(x,y)$ and applies non-maximum suppression to thin areas in the response map, where the responses are non-zero, to one pixel wide candidate points belonging to curvilinear structures.


The hysteresis thresholding requires a low and a high threshold parameter values, denoted by $t_l$ and $t_h$, respectively. We set $t_l=0.5t_h$ according to~\cite{grigorescu2004contour}. The resulting binary image depends on the given high threshold $t_h$: the lower that value the more line pixels in the binary image as less responses are suppressed. In Fig.~\ref{fig2}d we show the thinned and binarized response map of the proposed curved line operator applied to the image in Fig.~\ref{fig2}a.

We finally perform a morphological closing operation, with a $3 \times 3$ square structuring element, so as to fill eventual small gaps in the detected lines.

\section{Experiments}
\label{sec:results}


\begin{figure}[!t]
	\centering
	\footnotesize
   
   \subfloat[]{\label{fig:original}
	\includegraphics[width=27mm]{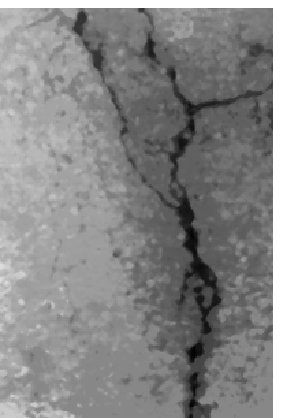}}
	   \subfloat[]{\label{fig:gt}
	\includegraphics[width=27mm]{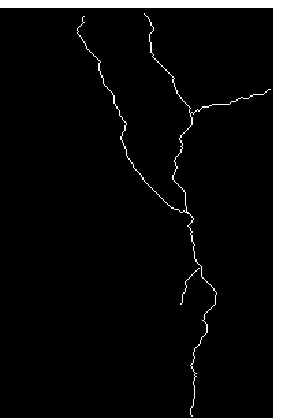}}
	   \subfloat[]{\label{fig:resp}
	\includegraphics[width=27mm]{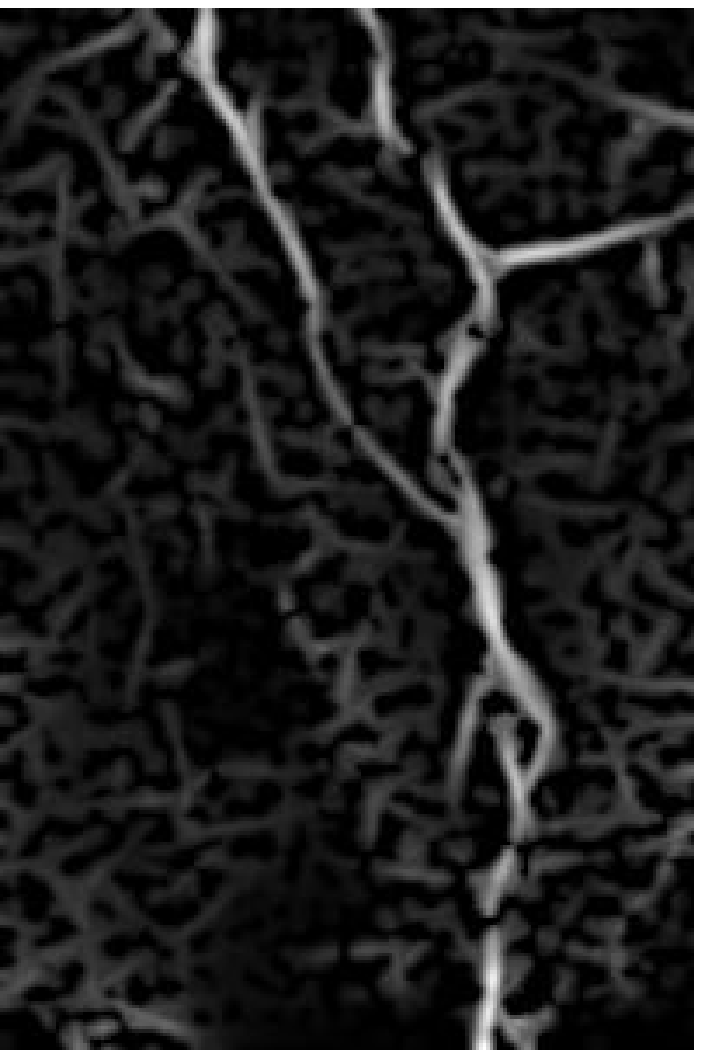}}
	   \subfloat[]{\label{fig:thinned}
	\includegraphics[width=27mm]{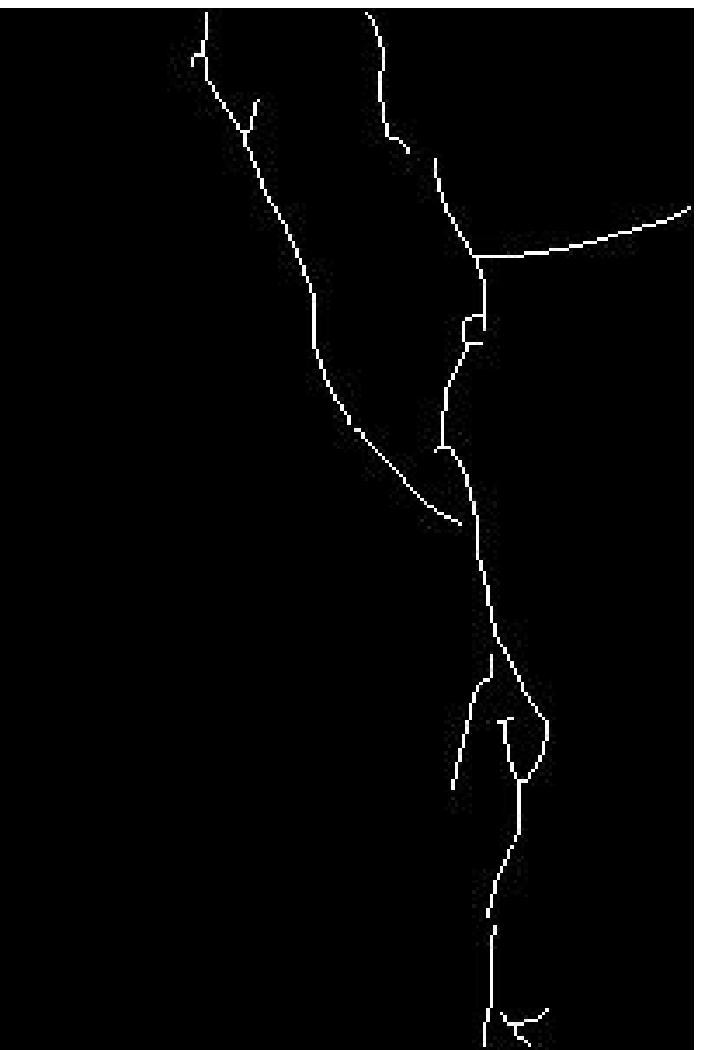}}
   \caption{(a) Example image of a pavement crack with (b) its corresponding manually annotated ground truth. (c) The response of a rotation-tolerant \textit{B}-COSFIRE filter and (d) the final thinned and binarized output image. }
   \label{fig:examples}
\end{figure}

\subsection{Data set}
We carried out experiments on a publicly available data set of pavement images called Crack\_PV14~\cite{Zou16}. This data set is composed of $14$ images taken with a laser range imaging appliance, mounted on the back of a car. We show an example image in Fig.~\ref{fig:original}. The images are distributed in BMP format and have resolution of $200 \times 300$ pixels. Each image is provided together with a manually annotated image that serves as ground truth for performance evaluation, Fig.~\ref{fig:gt}. The ground truth annotation for each image is a one-pixel wide line-network that delineates the center-line of the cracks.

As an example, we show the response of a rotation-tolerant \textit{B}-COSFIRE filter applied to the image in Fig.~\ref{fig:original} and its thinned an binarized version in Figs.~\ref{fig:resp} and~\ref{fig:thinned}, respectively.

\subsection{Evaluation}
In order to assess the performance of the proposed method and compare it with those of other approaches, we compute the precision (Pr), recall (Re) and F-measure (F) for each image in the Crack\_PV14 data set, as follows:
\begin{equation}
Pr = \frac{TP}{TP+FP}, Re = \frac{TP}{TP+FN}, F = \frac{2\cdot Pr \cdot Re}{Pr + Re},
\end{equation}
\noindent where TP are true positive pixels, FP are false positives and FN are false negatives. 
For each image we compute these three measurements for values of the threshold  $t_h$ from $0$ to $1$ in steps of $0.01$. Then, we compute the average F-measure $\bar{F}$ for each threshold value on the whole data set and choose the value of $t_h$ that contributes to the highest  $\bar{F}$ value.

According to~\cite{Zou16}, we consider some tolerance when computing the performance measures to compensate for some imprecision in the ground truth.
If the Euclidean distance $d$ of a detected crack point to the nearest crack point in the ground truth is lower than a value $d^*$, we consider that point as a true positive, otherwise it is a false positive. The points of the ground truth that are not detected within a distance $d^*$ in the output image are considered false negatives. As suggested in~\cite{Zou16} we set $d^* = 2$. Furthermore, we evaluate the overall performance of the proposed approach by plotting the Precision-Recall curve. This curve shows the trade-off between the Precision and Recall metrics as the value $t_h$ for the hysteresis thresholding varies.

\subsection{Results and discussion}
\begin{figure}[!t]
	\centering
	\footnotesize
   \input{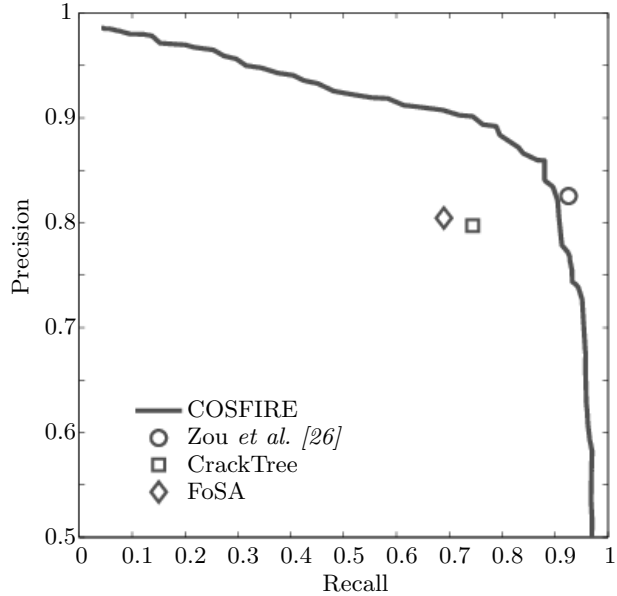}
   \caption{Precision-Recall curve achieved by the proposed approach on the Crack\_PV14 data set. The point $\bm{\lozenge}$  corresponds to the results of the FoSA method ($Pr = 0.8045$, $Re = 0.6896$), the point $\bm{\square}$  to the results of the CrackTree method ($Pr = 0.7972$, $Re = 0.7441$), while the point $\bm{\circ}$  to the ones of Zou \emph{et al.}~\cite{Zou16} ($Pr = 0.8254$, $Re = 0.9253$).}
   \label{fig:pr-re}
\end{figure}

Using this approach, we obtained an average F-measure $\overline{F}$ on the Crack\_PV14 data set equals to $0.865$ (with a standard deviation of $0.0975$). In Fig.~\ref{fig:pr-re} we plot the Precision-Recall curve obtained by the proposed method. On the same plot we indicate the points that correspond to the results achieved by other methods. The points corresponding to the CrackTree~\cite{Zou2012} and FoSA~\cite{Li11} approaches are considerably below our curve, and hence they are much less effective than our method. The point that represents the average results reported in~\cite{Zou16} is slightly above the Precision-Recall curve, which may indicate a better performance than our method. 
In order to clarify these indications, we evaluated the statistical significance of the results that we obtained with respect to the ones achieved by other methods by means of a paired $t$-test statistic. It turns out that there is significant statistical difference between the results of our method and those obtained by CrackTree and FoSA, but no statistical difference with respect to the ones obtained by Zou \emph{et al.}~\cite{Zou16}. Although we achieved comparable results with the ones obtained by the method of Zou \emph{et al.}, the proposed approach is based on a general algorithm for delineation of curvilinear structures in images, while other approaches are designed to solve a specific problem.

\begin{table}[!t]
\centering
\scriptsize
\renewcommand{\arraystretch}{1.3}
\begin{tabular}{c|C{25mm}C{20mm}C{25mm}C{15mm}}
\multicolumn{1}{c}{~} & \multicolumn{4}{c}{\bfseries F-Measure per image in the Crack\_PV14 data set} \\ \hline \hline
\bfseries Image & \bfseries Ours & \bfseries Zou~\emph{et al.}~\cite{Zou16} & \bfseries CrackTree~\cite{Zou2012} & \bfseries FoSA~\cite{Li11} \\  \hline 
$1$ & $0.899$ & $\bm{0.916}$ & $0.751$ & $0.721$ \\
$2$ & $0.867$ & $\bm{0.872}$ & $0.614$ & $0.64$ \\
$3$ & $\bm{0.97}$ & $0.874$ & $0.79$ & $0.728$ \\
$4$ & $0.822$ & $\bm{0.845}$ & $0.764$ & $0.722$ \\
$5$ & $0.886$ & $\bm{0.944}$ & $0.70$ & $0.691$ \\
$6$ & $\bm{0.95}$ & $0.747$ & $0.708$ & $0.729$ \\
$7$ & $0.818$ & $\bm{0.937}$ & $0.648$ & $0.623$ \\
$8$ & $0.836$ & $\bm{0.886}$ & $0.682$ & $0.698$ \\
$9$ & $0.69$ & $\bm{0.867}$ & $0.695$ & $0.70$  \\
$10$ & $\bm{0.99}$ & $0.892$ & $0.892$ & $0.808$ \\
$11$ & $0.668$ & $\bm{0.906}$ & $0.898$ & $0.858$ \\
$12$ & $0.854$ & $\bm{0.893}$ & $0.883$ & $0.835$ \\
$13$ & $\bm{0.872}$ & $0.639$ & $0.739$ & $0.679$ \\
$14$ & $\bm{0.982}$ & $0.913$ & $0.966$ & $0.901$ \\ \hline
$\overline{F}$ & $0.865$ & $\bm{0.866}$ & $0.766$ & $0.738$ \\
$\sigma_F$ & $0.0975$ & $\bm{0.0811}$ & $0.1057$ & $0.0821$ \\ \hline \hline
\multicolumn{5}{c}{\bfseries Results of paired \textit{t}-test statistic  } \\ \hline
$h$ & - & $0$ & $\bm{1}$ & $\bm{1}$ \\
$p$-value & - & $0.9608$ & $0.0131$ & $0.0015$ \\ \hline \hline
\end{tabular}
\vspace{3mm}
\caption{Detailed F-measure values achieved by the proposed approach in comparison with the ones obtained by other methods. The statistical significance of the F-measure difference with other methods is evaluated with a paired $t$-test statistic ($h=0$ indicates that the difference is not statistically significant while $h=1$ statistical significance). }
\label{tab:results}
\end{table}

In Table~\ref{tab:results} we report the F-measure that we obtained for each image in the Crack\_PV14 data set and compare them with the ones obtained by other methods. The results that we report are obtained by setting the hysteresis threshold $t_h$ equal to $0.49$, the one that contributed to the best overall results. 



The curved line operator based on \textit{B}-COSFIRE filters that we propose can be employed in image processing pipelines that require the delineation of elongated structures. In this work, we demonstrated the effectiveness of the proposed operator in the application of crack detection in images with noisy pavements.

The configuration parameters of a \textit{B}-COSFIRE filter determine its selectivity for lines of given width, length and orientation. For our experiments, we chose these values in a way that the configured filter is selective for average characteristics of the patterns of interest (i.e. the cracks) in the application at hand. We configured a single \textit{B}-COSFIRE filter with the following parameters: $w=6.34$, $l=29$, $\eta=2$, $\sigma_0=2$ and $\alpha=1$, which we determined by a grid search on 50\% of the images in the Crack\_PV14 data set.

In contrast to existing approaches for line detection, the \textit{B}-COSFIRE filters are not restricted to the detection of elongated structures. They can be configured to be selective for any pattern of interest in an automatic configuration step that is performed on a given prototype pattern. This possibility is a kind of \emph{representation learning}, which involves the construction of a data representation learned directly from training data. 
Similarly to deep learning and in contrast with traditional pattern recognition approaches, the COSFIRE approach avoids engineering of hand-crafted features and allows for the construction of flexible and adaptive pattern recognition systems.

One can configure filters that are selective for curvilinear structures of different sizes or shapes and combine their responses in order to improve the performance of the concerned method. As an example, in~\cite{AzzopardiMEDIA2015} we demonstrated how the responses of a \textit{B}-COSFIRE filter selective for vessels and one for vessel-endings were combined to improve the quality of the vessel delineation. Alternatively, machine learning techniques proposed in~\cite{Strisciuglio15,Strisciuglio2016},  based on genetic algorithms and learning vector quantization, can be utilized to select and combine the responses of the most discriminant filters from a large set of pre-configured ones. 

One of the strengths of the \textit{B}-COSFIRE filter approach is the tolerance it uses in its application phase, which is controlled by the parameters $\sigma_0$ and $\alpha$. 
This accounts for generalization capabilities and robustness  with respect to variations of the patterns of interest. 

The characteristics of the \textit{B}-COSFIRE filters, namely the possibility of combining the responses of filters selective for structures with different characteristics,  the automatic configuration step, which is explained in detail in \cite{AzzopardiMEDIA2015}, and the tolerance introduced in their application phase, make them a flexible tool for image processing and pattern recognition.
The \textit{B}-COSFIRE filters can be used for the design and implementation of systems which can be easily adapted and applied to various problems.

For the processing of a single image in the Crack\_PV14 data set (of $200 \times 300$ pixels),  our straightforward Matlab implementation takes an average time of $0.3575$ seconds (with a standard deviation of $0.0228$) on a machine with a $2.7$ GHz processor. The processing of a \textit{B}-COSFIRE filter is, however, parallelizable and hence the efficiency of the proposed approach can be further improved. 

\begin{figure}[!t]
\centering
\footnotesize
\setlength{\unitlength}{22mm}%
\begin{tabular}{cccc}
\includegraphics[height=\unitlength]{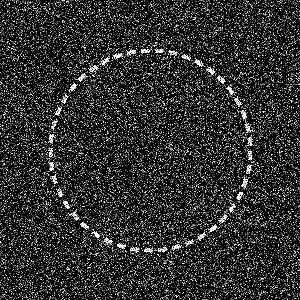} &
\includegraphics[height=\unitlength]{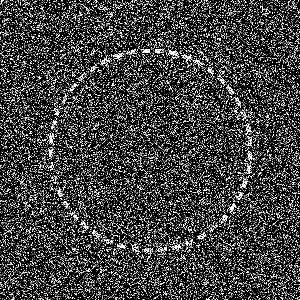} &
\includegraphics[height=\unitlength]{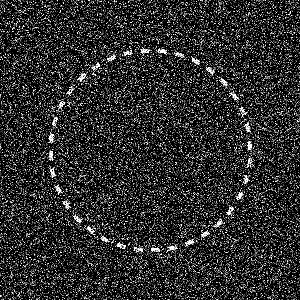} &
\includegraphics[height=\unitlength]{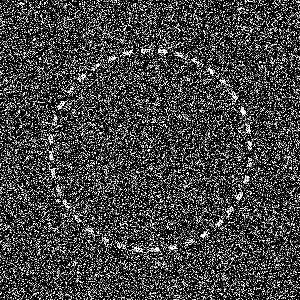} \\
(a) & (b) & (c) & (d) \\
\includegraphics[height=\unitlength]{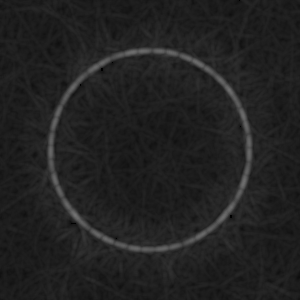} &
\includegraphics[height=\unitlength]{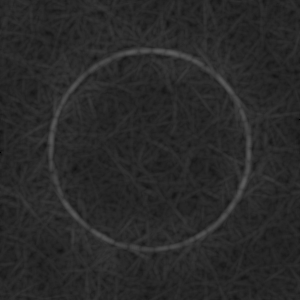} &
\includegraphics[height=\unitlength]{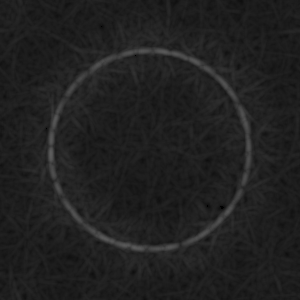} &
\includegraphics[height=\unitlength]{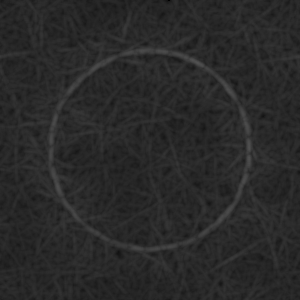} \\
(e) SNR=8.45 & (f) SNR = 4.60 & (g) SNR = 6.53& (h) SNR = 3.56\\
\includegraphics[height=\unitlength]{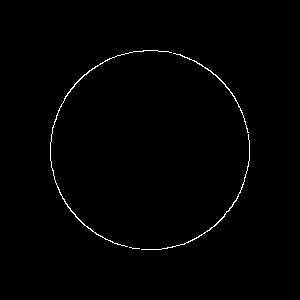} &
\includegraphics[height=\unitlength]{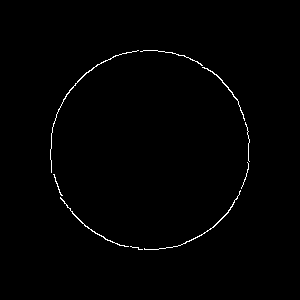} &
\includegraphics[height=\unitlength]{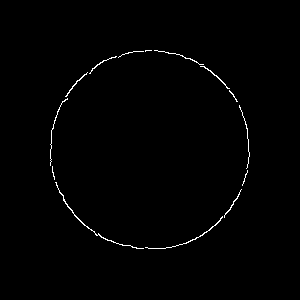} &
\includegraphics[height=\unitlength]{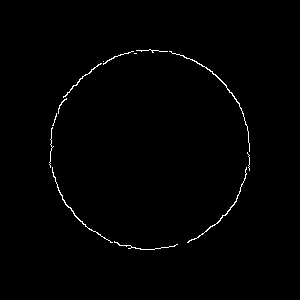} \\
(i) & (j) & (k) & (l) \\
\end{tabular}
\caption{(First row) Four stimuli (of size $300 \times 300$ pixels) of dashed circles with radii of 100 pixels and line width of 5 pixels, together with added Gaussian noise. The small arcs along the circumference are separated by (a,b) $3^\circ$ and (c,d) $5^\circ$ and the Gaussian noise in each image has zero mean and a variance of 0.2 in (a,c) and 0.5 in (b,d). (Second row) The corresponding rotation-tolerant response maps of the \textit{B}-COSFIRE filter. (Third row) The final thinned and binarized output maps with $t_h=0.75$.}
\label{fig3}
\end{figure}
\subsection{Robustness to noise, incomplete lines and tortuosity}
Fig.~\ref{fig3} provides insights about the robustness of the proposed method to noise and incomplete lines. In the first row we use four stimuli with dashed circles added with Gaussian noise of different variance. For each response map of the rotation-tolerant \textit{B}-COSFIRE filter (second row) we compute the signal-to-noise ratio ${SNR = 20\log_{10}(A_{s} / A_{n})}$, where $A_{s}$ is the average of all responses along the circumference of the circle in the input image, and $A_{n}$ is the average of all the other responses. The thinned and binarized results presented in the last row demonstrate the robustness of the proposed method.

\begin{figure}[!t]
\centering
\footnotesize
\setlength{\unitlength}{22mm}%
\begin{tabular}{ccc}
\includegraphics[height=\unitlength]{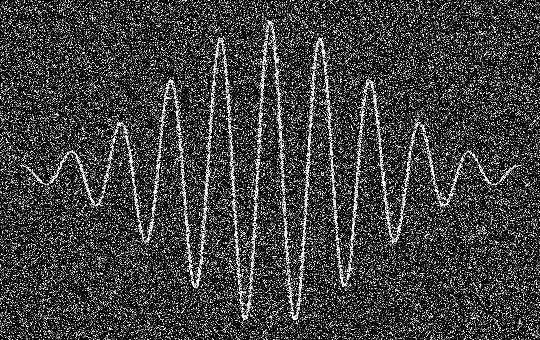} &
\includegraphics[height=\unitlength]{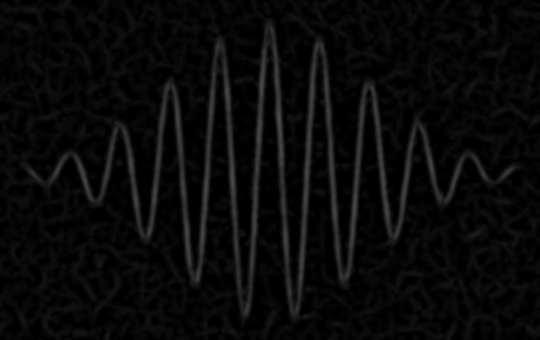} &
\includegraphics[height=\unitlength]{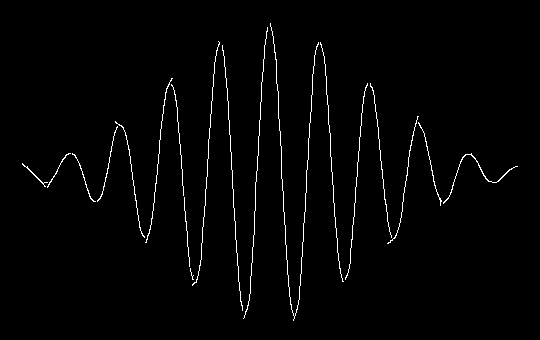} \\
(a) & (b) & (c) \\
\end{tabular}
\caption{(a) A synthetic image (of size $500 \times 300$ pixels) with Gaussian noise (zero mean and variance of $0.2$) superimposing a curvilinear structure that follows a one-dimensional Gabor function. (b) The response map of a \textit{B}-COSFIRE filter and (c) its thinned and binarized output ($t_h=0.75$).}
\label{fig:gabor}
\end{figure}

We also demonstrate the robustness of the proposed method with respect to tortuosity. In Fig.~\ref{fig:gabor}a we illustrate a stimulus with different degrees of tortuosity surrounded with Gaussian noise and in Fig.~\ref{fig:gabor}(b-c) we show the response map of the \textit{B}-COSFIRE filter and the final binary output.


\section{Conclusions}
\label{sec:conclusions}
The detector of curvilinear and elongated structures that we propose is highly effective in noisy images. We achieve state-of-the-art results (F-measure equals to $0.865$) on the Crack\_PV14  benchmark data set of images with noisy and cracked pavements. The proposed method is also very robust to incomplete lines and to linear structures with high tortuosity. The operator can be incorporated in computer vision applications that require delineation of curvilinear structures.

\bibliographystyle{splncs03}
\bibliography{citations}
\end{document}